%% file: main.tex
\documentclass{article}

\usepackage[preprint,nonatbib]{neurips_2020}
\input{packages}

\title{Noisy Agents: Self-supervised Exploration by Predicting Auditory Events}

\author{Chuang Gan$^{1,2}$\thanks{indicates equal contributions.}, \quad Xiaoyu Chen$^{3}$\footnotemark[1],\quad Phillip Isola$^{1}$, \\ \textbf{Antonio Torralba$^{1}$, \quad Joshua B. Tenenbaum$^{1}$} \\ \\
 $^1$ MIT,
 $^2$ MIT-IBM Watson AI Lab,
 $^3$ Tsinghua University
 \\\\
 \url{http://noisy-agent.csail.mit.edu/}}

\begin{document}

\maketitle

\begin{abstract}
Humans integrate multiple sensory modalities (\textit{e.g.}, visual and audio) to build a causal understanding of the physical world. In this work, we propose a novel type of intrinsic motivation for Reinforcement Learning (RL) that encourages the agent to understand the causal effect of its actions through auditory event prediction. First, we allow the agent to collect a small amount of acoustic data and use K-means to discover underlying auditory event clusters. We then train a neural network to predict the auditory events and use the prediction errors as intrinsic rewards to guide RL exploration. Experimental results on Atari games show that our new intrinsic motivation significantly outperforms several state-of-the-art baselines. We further visualize our noisy agents' behavior in a physics environment. We demonstrate that our newly designed intrinsic reward leads to the emergence of physical interaction behaviors (\textit{e.g.} contact with objects).


\end{abstract}

\section{Introduction}

Deep Reinforcement Learning algorithms aim to learn a policy of an agent to maximize its cumulative rewards by interacting with environments and have demonstrated substantial success in a wide range of application domains, such as video game ~\cite{mnih2015human}, board games~\cite{silver2016mastering}, and visual
navigation~\cite{zhu2017target}. While these results are remarkable, one of the critical constraints is the prerequisite of carefully engineered dense reward signals,  which are not always accessible. To overcome these constraints, researchers have proposed a range of intrinsic reward function. For example, curiosity-driven intrinsic reward based on prediction error of current~\cite{burda2018exploration} or future state\cite{pathak2017curiosity} on the latent feature spaces have shown promising results. Nevertheless, visual state prediction is a non-trivial problem as visual state is high-dimensional and tends to be highly stochastic in real-world environments.

The occurrence of physical events (\textit{e.g.} objects coming into contact with each other, or changing state) often correlates with both visual and auditory signals. Both sensory modalities should thus offer useful cues to agents learning how to act in the world.  Indeed, classic experiments in cognitive and developmental psychology show that humans naturally attend to both visual and auditory cues, and their temporal coincidence, to arrive at a rich understanding of physical events and human activity such as speech ~\cite{spelke1976infants,mcgurk1976hearing}. In artificial intelligence, however, much more attention has been paid to the ways visual signals (e.g., patterns in pixels) can drive learning.  We believe this misses important structure learners could exploit. As compared to visual cues, sounds are often more directly or easily observable causal effects of actions and interactions.  This is clearly true when agents interact: most communication uses speech or other nonverbal but audible signals.  However, it is just as much in physics.  Almost any time two objects collide, rub or slide against each other, or touch in any way, they make a sound. That sound is often clearly distinct from background auditory textures, localized in both time and spectral properties, hence relatively easy to detect and identify; in contrast, specific visual events can be much harder to separate from all the ways high-dimensional pixel inputs are changing over the course of  a scene.  The sounds that result from object interactions also allow us to estimate underlying causally relevant variables, such as material properties (\textit{e.g.}, whether objects are hard or soft, solid or hollow, smooth, or rough), which can be critical for planning actions.  

These facts bring a natural question of how to use audio signals to benefit policy learning in RL. In this paper, our main idea is to use sound prediction as an intrinsic reward to guide RL exploration. Intuitively, we want to exploit the fact that sounds are frequently made when objects interact, or other causally significant events occur, like cues to causal structure or candidate subgoals an agent could discover and aim for. A na\"ive strategy would be to directly regress feature embeddings of audio clips and use feature prediction errors as intrinsic rewards. However, prediction errors on feature space do not accurately reflect how well the agent understands the underlying causal structure of events and goals. It also remains an open problem on how to perform appropriate normalizations to solve intrinsic reward diminishing issues.  To bypass these limitations, we formulate the sound-prediction task as a classification problem, in which we train a neural network to predict auditory events that occurred after applying action to a visual scene. We use classification errors as an exploration bonus for deep reinforcement learning. Concretely, our pipeline consists of two exploration phases. In the beginning, the agent receives an incentive to actively collect a small amount of auditory data by interacting with the environment. Then we cluster the sound data into auditory events using K-means.  In the second phase,  we train a neural network to predict the auditory events conditioned on the embedding of visual observations and actions. The state that has the wrong prediction is rewarded and encouraged to be visited more. We demonstrate the effectiveness of our intrinsic motivation module on 25 Atari Games and a rolling robot multi-modal physic simulation platform build on top of TDW~\cite{gan2020threedworld}. In summary, our work makes the following contributions:
\setdefaultleftmargin{1em}{1em}{}{}{}{}
\begin{compactitem}

    \item We introduce a novel and effective auditory event prediction (AEP) framework to make use of the auditory signals as intrinsic rewards for RL exploration. 

    \item Our system outperforms previous state-of-the-art vision only curiosity-driven exploration agents on most of the Atari games.
    
    \item We show that our new intrinsic module is more stable in the 3D multi-modal physical world environment and can encourage interest actions that involved physical interactions. 
\end{compactitem}

\section{Related Work}

\textbf{Audio-Visual Learning.} In recent years, audio-visual learning has been studied extensively. By leveraging audio-visual correspondences in videos, it can help to learn powerful audio and visual representations through self-supervised learning~\cite{owens2016ambient,aytar2016soundnet,arandjelovic2017look,korbar2018co,owens2018audio}. Other interesting applications using audio-visual knowledge transfer include sounding object localization\cite{senocak2018learning,arandjelovic2018objects}, sound source separation~\cite{gao2018object-sounds,gan2020music,Zhao_2018_ECCV,ephrat2018looking,zhao2019sound,afouras2018conversation}, biometric matching~\cite{nagrani2018seeing}, sound generation for videos \cite{owens2016visually,zhou2017visual,gao20182,morgado2018self,gan2020Foley}, audio-visual co-segmentation~\cite{rouditchenko2019self}, auditory vehicle tracking~\cite{gan2019self} and action recognition~\cite{long2018attention,long2018multimodal,nagrani2020speech2action}. In contrast to the widely used correspondences between these two modalities, we take a step further by considering sound as causal effects of actions.

\textbf{RL Explorations.} The problem of exploration in Reinforcement Learning (RL) has been an active research topic for decades. There are various solutions that have been investigated for encouraging the agent to explore novel states, including rewarding information gain~\cite{little2013learning}, surprise~\cite{schmidhuber1991possibility,schmidhuber2010formal},  state visitation counts~\cite{tang2017exploration,bellemare2016unifying}, empowerment~\cite{klyubin2005empowerment}, curiosity~\cite{pathak2017curiosity,burda2018large} disagreement~\cite{pathak2019self} and so on. A separate line of work~\cite{osband2017deep,osband2016deep} studies adopt parameter noises and Tompson sampling heuristics for exploration. For example, Osband \cite{osband2017deep} trains multiple value functions and makes use of the bootstraps for deep exploration. Here, we mainly focus on the problem of using intrinsic rewards to drive explorations. The most widely used intrinsic motivation could be roughly divided into two families. The first one is count-based approaches~\cite{strehl2008analysis,bellemare2016unifying,tang2017exploration,ostrovski2017count,martin2017count,burda2018exploration}, which encourage the agent to visit novel states. For example, Burda \cite{burda2018exploration} employs the prediction errors of a self-state feature extracted from a fixed and random initialized network as exploration bonuses and encourage the agent to visit more previous unseen states. Another one is the curiosity-based approach~\cite{stadie2015incentivizing,pathak2017curiosity,haber2018learning,burda2018large}, which is formulated as the uncertainty in predicting the consequences of the agent's actions. For instance, \cite{pathak2017curiosity,burda2018large} uses the errors of predicting the next state in the latent feature space as rewards. The agent is then encouraged to improve its knowledge about the environment dynamics. In contrast to previous work that purely works on visual observations, we make use of the sound signals as rewards for RL explorations.

\textbf{Sounds and Actions.} There are numerous works to explore the associations between sounds and actions. For example, Owens \cite{owens2016visually} made the first attempt to collect an audio-video dataset through physical interaction with objects and train an RNN model to generate sounds  for silent videos. \cite{shlizerman2018audio,ginosar2019learning} explore the problem of predicting body dynamics from music and body gesture from speech.  Gan ~\cite{gan2019look} and Chen ~\cite{chen2019audio} introduce an interesting audio-visual embodied task in 3D simulation environments.
More recently, Gandhi~\cite{Gandhi20} collected a large sound-action-vision dataset using Tilt-bolt and demonstrates  sound signals could provide valuable information for find-grained object recognition, inverse model learning, and forward dynamic model prediction. More related to us are the papers from \cite{aytar2018playing} and \cite{omidshafiei2018crossmodal}, which have shown that the sound signals could provide useful supervisions for imitation learning and reinforcement learning in Atari games. Concurrent to our work, \cite{dean2020see} uses novel associations of audio and visual signals as intrinsic rewards to guide RL exploration. Different from them, we mainly studied if the sound signals alone could be utilized as intrinsic rewards for RL explorations.

\section{Method}
In this section, we first introduce some background knowledge of reinforcement learning and intrinsic rewards. Then we will present the representations of auditory events. Finally, we elaborate on the pipeline of self-supervised exploration through auditory event predictions. The pipeline of our system is outlined in Figure~\ref{fig:framework}.

\input{teaser.tex}
\subsection{Background}

\newcommand{\MDP}{\Gamma}
\newcommand{\MDPS}{\mathcal{S}}
\newcommand{\MDPA}{\mathcal{A}}
\newcommand{\MDPT}{\mathcal{T}}
\newcommand{\MDPR}{\mathcal{R}}
\newcommand{\MDPDIS}{\gamma}
\newcommand{\MDPPOLICY}{\pi _ {\MDPA}}
\newcommand{\MRDPL}{\mathcal{L}}
\newcommand{\MRDPPOLICY}{\pi _ {\MRDPL}}
\noindent\textbf{MDPs} We formalize the decision procedure in our context as a standard Markov Decision Process (MDP), defined as  $(\mathcal{S}, \mathcal{A}, r, \mathcal{T}, \mu, \gamma)$.  $\mathcal{S}$, $\mathcal{A}$ and $\mu(s): S \rightarrow [0, 1]$ denote the sets of state, action and the distribution of initial state respectively. The transition function $\mathcal{T}(s'|s,a): \mathcal{S} \times \mathcal{A} \times \mathcal{S} \rightarrow [0, 1]$ defines the transition probability to next-step state $s'$ if the agent takes action $a$ at current state $s$.
The agent will receive a reward $r$ after taking an action $a$ according to the reward function $\mathcal{R}(s, a)$, discounted by $\gamma \in (0, 1)$.
The goal of training reinforcement learning is to learn an optimal policy $\pi^{*}$ that can maximize the expected rewards under the discount factor $\gamma$ as
\begin{equation}
   \pi^{*} =  \mathop{\arg\max}_{\pi} E_{\zeta \in \pi} \left \lceil \sum R(s,t) \gamma_{t} \right \rceil
\end{equation}
where $\zeta$ represents the agent's trajectory, namely $\{(s_0, a_0), (s_1, a_1), \cdots\}$. 
The agent chooses an action $a$ from a policy $\pi(a|s): \mathcal{S} \times \mathcal{A} \rightarrow [0,1]$ that specifies the probability of taking action $a\in \MDPA$ under state $s$. In this paper, we concentrate on the MDPs whose states are raw image-based observations as well as audio clips, actions are discrete, and $\MDPT$ is provided by the game engine.

\noindent\textbf{Intrinsic Rewards for Exploration.}  Designing intrinsic rewards for exploration has been widely used to resolve the sparse reward issues in the deep RL communities. One effective approach is to use the errors of a predictive model as exploration bonuses ~\cite{pathak2017curiosity,haber2018learning,burda2018large}. The intrinsic rewards will encourage the agent to explore those states with less familiarity. In this paper, we aim to train a policy that can maximize the errors of auditory event predictions.

\subsection{Representations of Auditory Events} 

Consider an agent that sees a visual observation $s_{v,t}$, takes an action $a_t$ and transits to the next state with visual observation $s_{v,t+1}$and sound effect $s_{s,t+1}$. The main objective of our intrinsic module is to predict auditory events of the next state, given feature representations of the current visual observation $s_{v,t}$ and the agent' action $a_t$. We hypothesize that the agents, through this process, could learn the underlying causal structure of the physical world and use that to make predictions about what will happen next, and as well as plan actions to achieve their goals.

To better capture the statistic of the raw auditory data, we extract sound textures~\cite{mcdermott2011sound} $\Phi(s_{s,t})$ to represent each audio clip $s_t$ $s_{s,t}$. For the task of auditory event predictions, perhaps the most straightforward option is to directly regress the sound features $\Phi(s_{s,t+1})$ given the feature embeddings of the image observation $s_{v, t}$and agent' actions $a_t$. Nevertheless, we find that not very effective. The reasons are mainly in two folds: 1) the sound textures do not explicitly capture the high-level events information; 2) the distances between the sound textures could not accurately reflect how well the agents grasp the underlying causal structure of these auditory events. For example, from the position of an aircraft and the shooting action, we hope the agent can infer that a critical event like an explosion will happen, rather than the intricate rhythm of the bang. Therefore, we choose instead to define explicit auditory events categories and formulate this auditory event prediction problem as a classification task, similar to~\cite{owens2016ambient}.

\input{events}
\subsection{Auditory Events Prediction Based Intrinsic Reward}

Our AEP framework consists of two stages: sound clustering and auditory event prediction. We need to collect a small set of diverse auditory data in the first stage and use them to define the underlying auditory event classes. To achieve this goal, we first train an RL policy that rewards the agents based on sound novelty. And then, we run a K-means algorithm to group these data into several auditory events. In the second phase, we train a forward dynamic network that takes input as the embedding of visual observation and action and predicts which auditory event will happen next. The prediction error is then utilized as an intrinsic reward to encourage the agent to explore those auditory events with more uncertainty, thus improving its ability to understand the consequences of its actions. We will elaborate on the details of these two-phase below.

\noindent\textbf{Sound clustering.} The agents start to collect audio data by interacting with the environment. The goal of this phase is to gather diverse data that could be used to define auditory events. For this purpose, we train an RL policy by maximizing the occurrences of novel sound effects. In particular, we design an online clustering-based intrinsic motivation module to guide explorations. Assuming we have a series of sound embeddings $\Phi(s_{s,t}), t \in {1, 2, ..., T}$ and temporarily grouped into $K$ clusters. Given a new coming sound embedding $\Phi(s_{s, t+1})$, we compute its distance to the closest cluster centers and use that as an exploration bonus. Formally,
\begin{equation}
    r_{t} = \min_{i} ||\Phi(s_{s, t+1}) - c_{i}||_2
\end{equation}
where $r_t$ denotes an intrinsic reward at time $t$, and   $c_{i}, i \in \{1, 2, ..., K\}$ represents a cluster center.
During this exploration, the number of clusters will grow, and each cluster's center will also be updated. Through this process, the agent is encouraged to collect novel auditory data that could enrich cluster diversity. After the number of the clusters is saturated, we then perform the K-Means clustering algorithm~\cite{Lloyd1982LeastSQ} on the collected data to define the auditory event classes and use the  center of each cluster for the subsequent auditory event prediction task. We visualize the corresponding visual states in two games (\textit{Frostbite} and \textit{Assault}) that belong to the same sound clusters, and it can be observed that each cluster always contains identical or similar auditory events (see Figure~\ref{fig:events})

\noindent\textbf{Auditory event predictions.} Since we have already explicitly defined the auditory event categories, the prediction problem can then be easily formulated as a classification task. We label each sound texture with the index of the closest center, and then train a forward dynamic  network $f(\Psi(s_{v,t}),a_{t}; \theta_p)$ that takes the embeddings of visual observation $\Psi(s_{v,t})$ and action $a_{t}$ as input to predict which auditory event cluster the incurred sound $\Phi(s_{s,t+1})$ belongs to. The forward dynamic model is trained on collected data using gradient descent to minimize the cross-entropy loss $L$ between the estimated class probabilities with the ground truth distributions $y_{t+1}$ as:
\begin{equation}
     L = {\rm Loss}(f(\Psi(s_{v,t}),a_{t}; \theta_p). y_{t+1})
\end{equation}
The prediction is expected to fail for novel associations of visual and audio data.  We will reward the agent at the that stage and encourage it to visit more since it is uncertain with about this scenario. In practice, we do find that the agent can learn to avoid dying scenario in the games since that gives a similar sound effect it has already encountered many times and can predict very well. By avoiding potential dangers and keeping seeking novel events, agents can learn causal knowledge of the world for planing their actions to achieve the goal.

\section{Experiments}
In the experiments, we investigate the following questions:
\begin{compactitem}

  \item Does the proposed audio-driven explorations outperform other intrinsic reward modules on learning skills without extrinsic rewards?
  \item Can our approach be combined with extrinsic rewards to improve policy learning on hard exploration Atari games?   
  \item What is the agent's behavior using different methods in a 3D multi-modality physical environment?
  \item Is each component in our methods necessary?
\end{compactitem}

\subsection{Setup}

\noindent\textbf{Atari Game Environment} Our primary goal is to investigate whether we could use auditory event prediction as intrinsic rewards to help RL exploration. For this purpose, we use the Gym Retro~\cite{nichol2018retro} as a testbed to measure agents' competency quantitatively. Gym Retro consists of a diverse set of Atari games, and also support an audio API to provide the sound effects of each state. We first use 20 familiar video games that contain the sound effects to compare the intrinsic reward only exploration against several previous state-of-the-art intrinsic modules. Then, we use five hard exploration games to investigate whether the newly designed motivation could be combined with extrinsic rewards to improve the results

\noindent\textbf{Rolling Robot Multi-Modality Simulation Platform.} We also test our module on a 3D multi-modal physic simulation platform. We build this platform on top of the Unity game engine and physics-based impact sound simulation toolbox from TDW~\cite{gan2020threedworld}. As shown in Figure~\ref{fig:physics} , we place a rolling robot agent (\textit{i.e.}, red sphere) on a billiard table. The agent can execute actions to interact with objects of different materials and shapes. When two objects collide, the environment could generate collision sound based on the physical properties of objects. We  would like to compare the agent's behaviors in this 3D world physical environment using different intrinsic rewards. 

\noindent\textbf{Implementation details.} For all the experiments, we choose PPO algorithm~\cite{schulman2017proximal}  based on the PyTorch implementation to train our RL agent since it is robust and requires little hyper-parameter tuning.  For experiments on Atari games, we use gray-scale image observations of size 84$\times$84 and  60ms audio clip. We set the skip frame  S$=$4 for all the experiments. We use a 4-layer CNN as the encoder of the policy network. As for the auditory prediction network, we choose a 3-layer CNN to encode the image observation and use 2-layer MLP to predict the auditory events.

\input{result_intrinsic.tex}
\subsection{Explorations without Extrinsic Rewards}

We first aim to compare how agents use different intrinsic rewards to explore the environment without extrinsic rewards. To quantify how well an exploration strategy, we use the external rewards it can achieve as an evaluation metric. It is important to note that the extrinsic reward is only used for evaluation, not for training.  We consider five state-of-the-arts intrinsic motivation modules for comparisons, including Intrinsic Curiosity Module (ICM)~\cite{pathak2017curiosity}, Random Feature Networks (RFN)~\cite{burda2018large}, Random Network Distillation (RND)~\cite{burda2018large}, and Model Disagreement (DIS)~\cite{pathak2019self}.  We run all the experiments for 10 million steps with 8 parallel environments.

Figure~\ref{fig:intrinsic} summarizes the evaluation curves of mean extrinsic reward in 20 Atari games. For each method, we experiment with three different random seeds and report the standard deviation in the shaded region. As the figure shows, our module achieves significantly better results than previous vision-only intrinsic motivation modules in 15 out of 20 games and is comparable in the other 5 games. Another interesting observation is that the earned score keeps going up, even only using intrinsic rewards. We observe that the agent could learn to avoid dangerous situations with dead sound, which might frequently happen at the beginning of explorations. There are also some failure cases in the above Atari games. For example, our method falls short on the games with background music or noises that could not reflect any auditory events (\textit{e.g.} Freeway and Time Pilot). Advanced audio processing algorithms might help, and we leave this to future work. \textbf{More 
model analysis and demo videos could be found in supplementary materials.}

\vspace{-3mm}
\subsection{Combining Extrinsic and Intrinsic Rewards for Hard Explorations}

The intrinsic rewards could serve as incentives that allow the agent to distinguish novel and fruitful states, but the lack of extrinsic rewards impedes the awareness of auditory events where agents can earn more rewards and need to visit again. In this section, we investigate whether the audio-driven intrinsic rewards could be further utilized to improve policy learning in the hard exploration scenarios, where extrinsic rewards exist but very sparse. We use five hard exploration environments in Atari games, including Venture, Solaris, Private Eye, Pitfall!, and Gravitar for experiments.  Following the strategy proposed by RND~\cite{burda2018large}, we use two value heads separately for the intrinsic and extrinsic reward module and then combine their returns. We also normalize intrinsic rewards to make up the variances among different environments. We compare our model against the plain PPO using purely extrinsic rewards. All experiments are run for 4 million steps with 32 parallel environments. We use the max episodic returns to measure the ability of explorations.

The comparison results are shown in Figure~\ref{fig:extrinsic}.
We find that our audio-driven exploration can significantly improve the policy learning of hard exploration games in Atari. For example, when combining the intrinsic reward, the agent can earn four times rewards on game Venture and almost 30 times rewards Private Eye.

\input{hard_exploration.tex}

\subsection{Understanding Explorations Behaviors in 3D Physical World}
In this section, we would like to see the agent's behaviors in a near photo-realistic 3D physical world. A curious rolling robot is required to use interactions to build causal models of the physical world.
\input{physics}

\noindent\textbf{Setup.} For this experiment, we take an image observation of 84$\times$84 size and 50ms audio clip as input. We use a three-layer convolutional network to encode the image and extract sound textures from the audio clip. Same as the previous experiment, we train the policy using the PPO algorithm. The action space consists of moving to eight directions and stop. The action is repeated 4 times on each frame. We run all the experiments for 200K steps with 8 parallel environments. 

\noindent\textbf{Result Analysis.} To understand and quantify the agent's behaviors in this environment, we show the number of collision events and intrinsic reward rewards in Figure~\ref{fig:physics}. We noticed two major issues with the previous vision-based curiosity model (\textit{e.g.} RND and ICM). First, the prediction errors on latent features space could not accurately reflect subtle object state changes in the 3D photo-realistic world, in which a physical event happens. Second,  the intrinsic reward can diminish quickly during training, since the learned predictive model usually converges to a stable state representation of the environment. Instead, our auditory event prediction driven exploration will lead agents to interact more with objects in the physical world, which is critical to learning the dynamics of the environments.

\subsection{Ablated Study}
In this section, we perform in-depth ablation studies to evaluate each component of our model using 3 Atari games with apparent sound effects: Amidar, Battle Zone, and Carnival. All experiments are run for 2 million steps with 8 parallel environments.

\noindent\textbf{Predict auditory events or sound features?} One main contribution of our paper is to use auditory event prediction as an intrinsic reward. To further understand this module's ability, we conduct an ablated study by replacing this module with sound feature prediction module. In particular, we train a neural network that takes the embedding of visual state and action as input and predicts the sound textures. The comparison curves are plotted in Figure~\ref{fig:event}. We observe that the auditory event prediction module indeed earned more rewards. This result demonstrates the advantage of using auditory event classes over latent sound feature embedding for RL explorations. We speculate that auditory events provide more structured knowledge of the world, thus lead to better policy learning. 

\noindent\textbf{Sound clustering or auditory event prediction?} We adopt a two-stage exploration strategy. A natural question is if this is necessary. We show the curve of using sound clustering only as an intrinsic reward in Figure~\ref{fig:event}.  We notice that the returned extrinsic reward is similar to sound feature prediction, but  worse than auditory event prediction.

\noindent\textbf{Active exploration or random explorations?} We propose an online clustering-based intrinsic module for active audio data collections. To verify its efficacy, we replace this module with random explorations. For fair comparisons, we allow both models to use 10K interaction data to define the event classes with the same K-mean clustering algorithm. The comparison results are shown in Figure~\ref{fig:active}.  We can find that the proposed active explorations indeed achieve better results. We also compute the cluster distances of both two models and find that the sound clusters discovered by active exploration are much diverse, thus facilitate the agents to perform in-depth explorations.

\input{ablated.tex}

\section{Conclusions}

In this work, we introduce an intrinsic reward function of predicting sound for RL exploration. Our model
employs the errors of auditory event prediction as an exploration bonuses, which allows RL agent to explore novel physical interactions of objects. We demonstrate our proposed methodology and compared it against a number of baselines on Atari games.  Based on the experimental result above, we therefore conclude that sound conveys rich information and is powerful for agents to build a causal model of the physical world for efficient RL explorations. We hope our work could inspire more works on using multi-modality cues for planing and control.

\section*{Broader Impact}
Our work is on the basic science of multimodal learning and exploration in RL. Since auditory signals are prevalent in real-world scenarios, we believe that combining them with visual signals could help guide exploration in many robotic applications~\cite{Gandhi20,gan2019look,chen2019audio}. For example, the honk of a car may be a useful signal that a self-driving agent has entered an unexpected situation. This example also raises a potential negative use of the ideas in this paper: if a self-driving car explores by seeking out honks it would likely put humans in danger. Future work should therefore consider how to combine the ideas in this paper with the notion of safe exploration.

Studying the role of auditory and visual signals could also be especially relevant for sight and hearing impaired populations. For example, if we better understand the role of audition in exploration, perhaps we can develop applications that better serve deaf users, who lack this signal. 

A limitation of our work is that it only experiments on synthetic environments, which may not reflect realistic scenarios. For example, Atari games have sound effects that often correlate with game achievements, whereas the correlation between sound and reward in nature is likely much more complex. The findings in our paper can therefore be considered to be biased by the design of the synthetic environments. Future work will be necessary to validate our methods on real world applications.


\medskip
\small

\bibliographystyle{plain}
\bibliography{egbib}
\input{supp}

\end{document}

%% file: packages.tex
\usepackage{color,xcolor}
\usepackage{epsfig}
\usepackage{graphicx}

\usepackage{adjustbox}
\usepackage{array}
\usepackage{booktabs}
\usepackage{colortbl}
\usepackage{wrapfig}
\usepackage{hhline}
\usepackage{multirow}
\usepackage{subcaption} 

\usepackage{amsmath,amsfonts,amssymb}
\usepackage{bm}
\usepackage{nicefrac}
\usepackage{microtype}

\usepackage{changepage}
\usepackage{extramarks}
\usepackage{fancyhdr}
\usepackage{lastpage}
\usepackage{setspace}
\usepackage{soul}
\usepackage{xspace}

\usepackage[pagebackref=true,breaklinks=true,colorlinks,citecolor=green]{hyperref}

\usepackage{url}

\usepackage{enumerate}
\usepackage{todonotes} 
\usepackage{enumitem}  
\usepackage{paralist}
\usepackage{titlesec}
\usepackage{verbatim}
\usepackage{makecell}

\usepackage{pifont} 

%% file: teaser.tex
\begin{figure*}
    \centering 
    \includegraphics[width=1\linewidth]{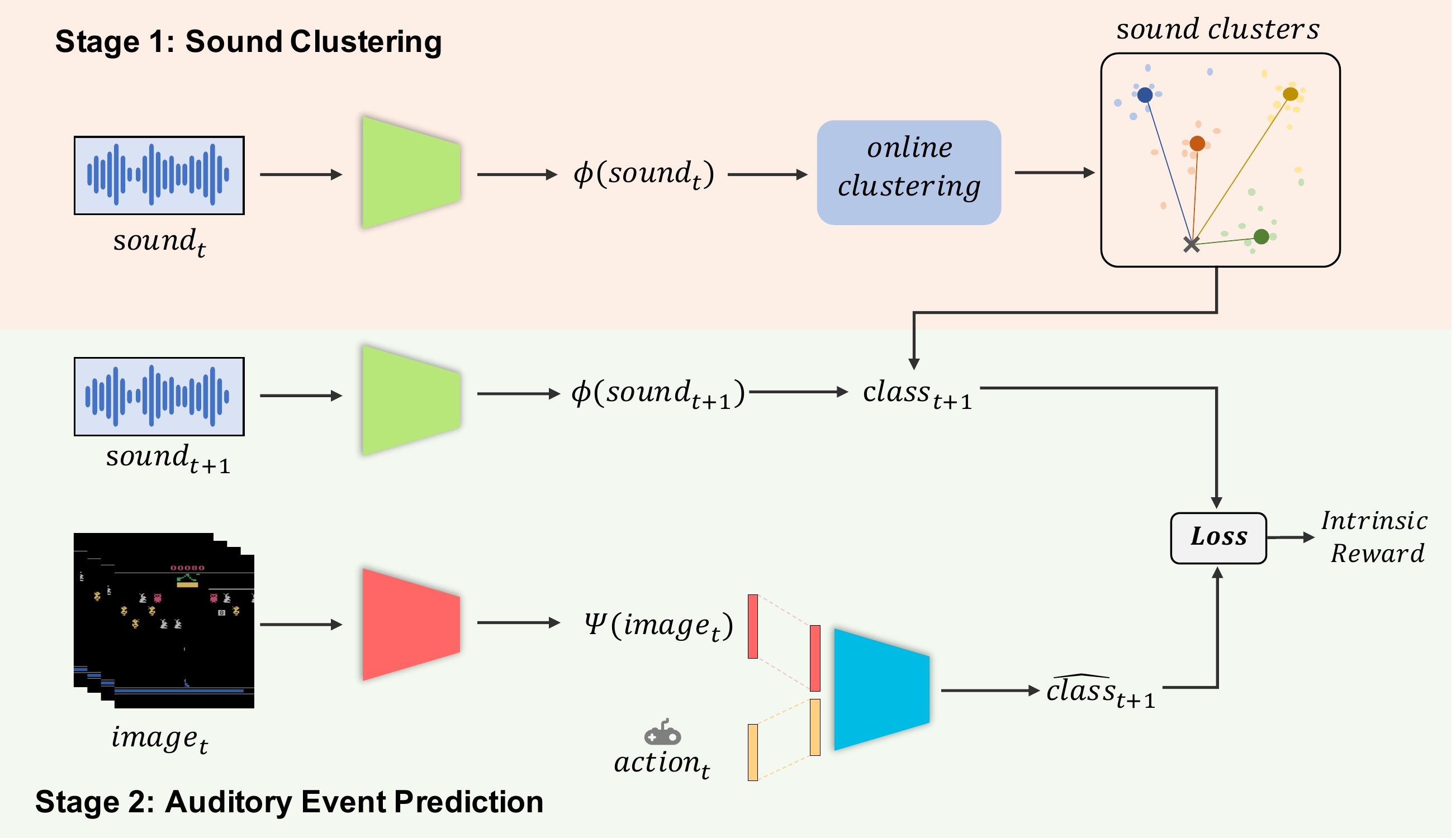}
    \caption{\textbf{An overview of our framework.} Our model consists of two stages: sound clustering and auditory event prediction. The agent start to collect a diverse set of sound through limited environment interactions (\textit{e.g.} 10K) and then cluster them into auditory event classes. In the second stage, the agent use errors of auditory events predictions as intrinsic reward to explore the environment.}
    \label{fig:framework}
    \vspace{-5mm}
\end{figure*}

%% file: events.tex
\begin{figure*}[t]
    \centering 
    \includegraphics[width=1\linewidth]{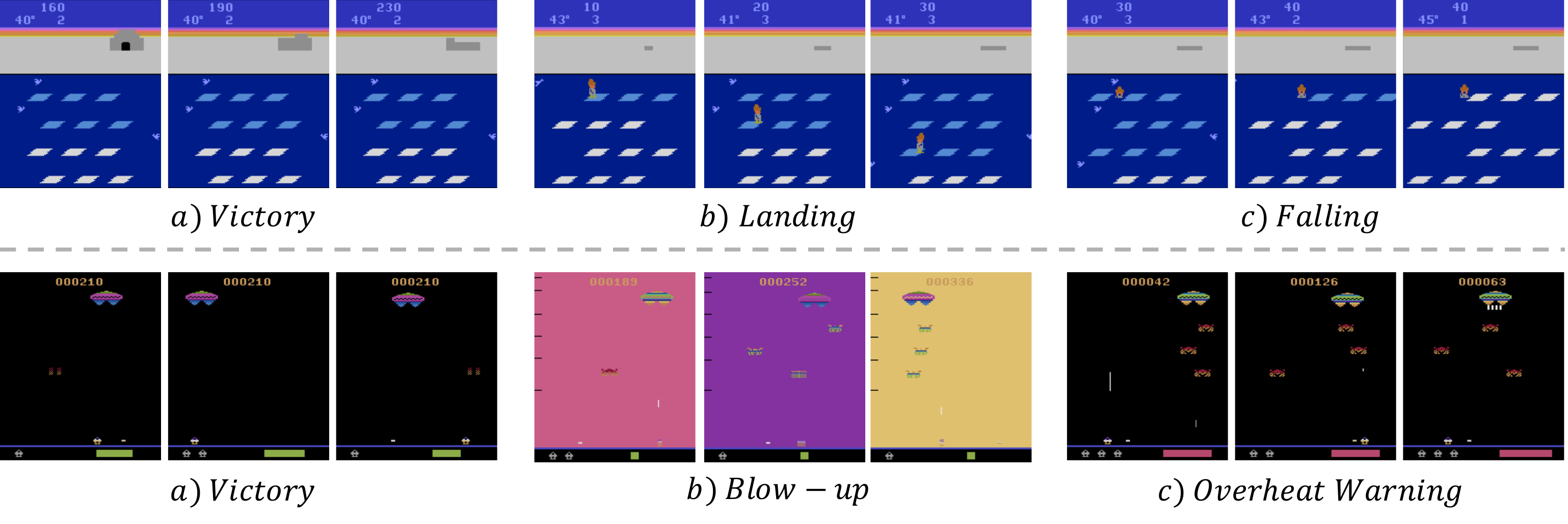}
       \vspace{-5mm}
    \caption{ The first and second rows show the auditory events that we discovered by the K-means algorithm in \textit{Frostbite} and \textit{Assault}.}
    \vspace{-5mm}
    \label{fig:events}
\end{figure*}

%% file: result_intrinsic.tex
\begin{figure*}
    \centering 
    \includegraphics[width=1\linewidth]{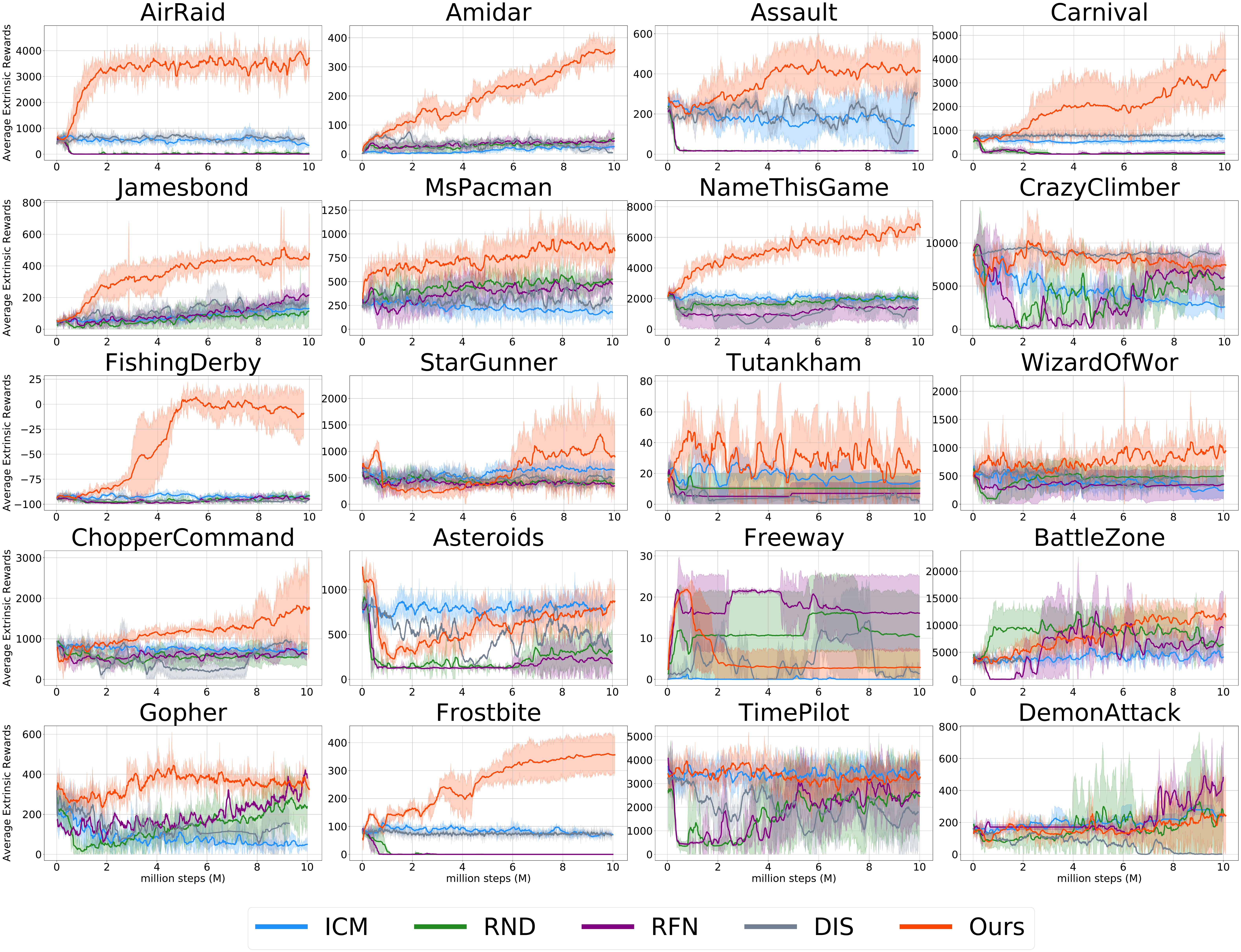}
    \caption{Average extrinsic rewards of our model against baselines in 20 Atari games. }
    \label{fig:intrinsic}
 
\end{figure*}

%% file: hard_exploration.tex
\begin{figure*}
    \centering 
    \includegraphics[width=1\linewidth]{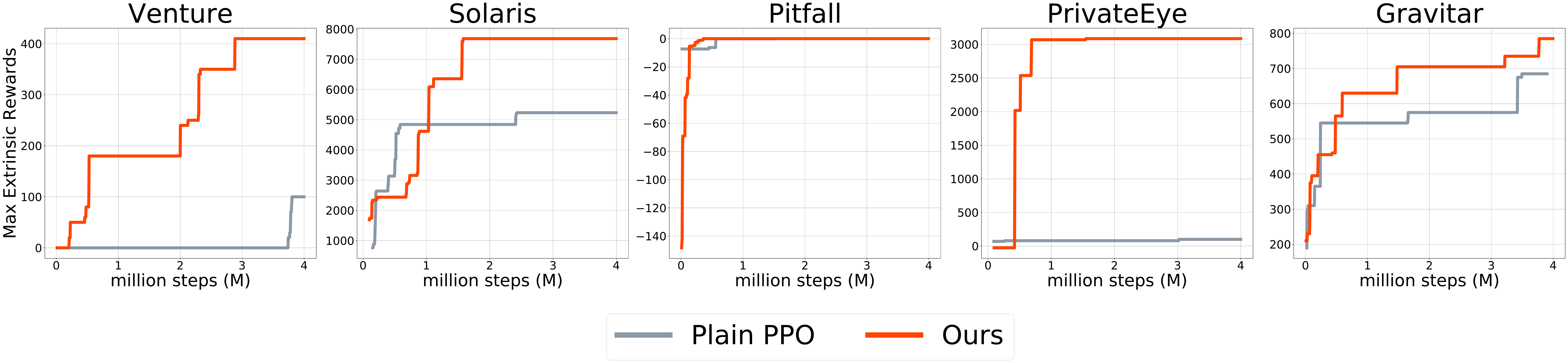}
     \vspace{-4mm}
    \caption{Comparison of combining intrinsic and extrinsic reward against using extrinsic reward only on 5 hard exploration Atari Games.}

    \label{fig:extrinsic}
\end{figure*}

%% file: physics.tex
\begin{figure*}
    \centering 
    \includegraphics[width=1\linewidth]{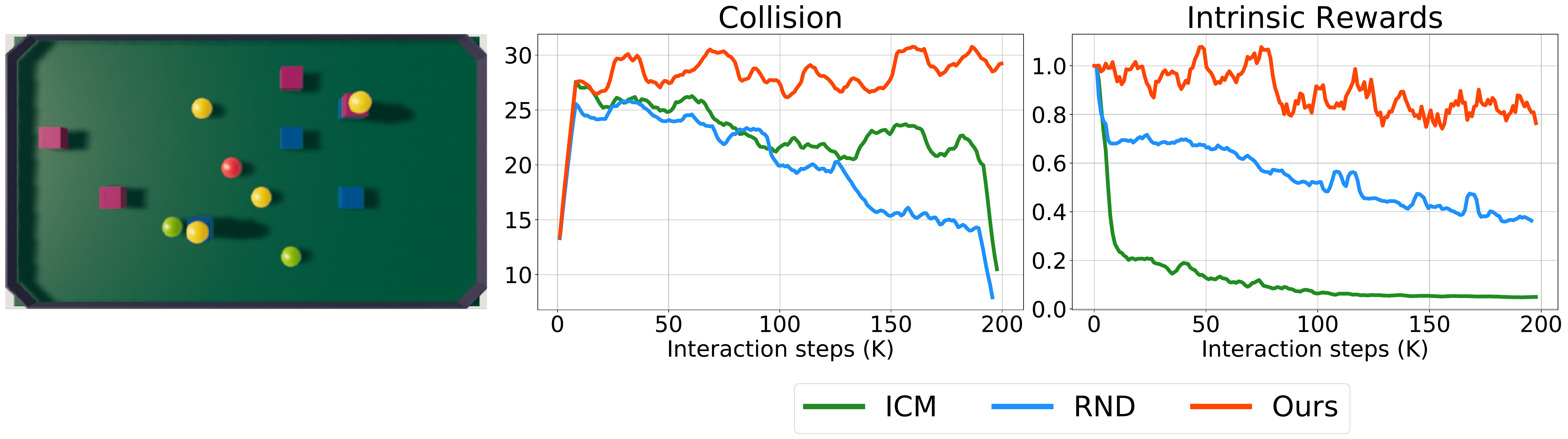}
    \caption{Explorations on a multi-modal physics environment. From left to right: physical scene, collision events, and intrinsic reward changes }
    \label{fig:physics}
\end{figure*}

%% file: ablated.tex
\begin{figure*}

\begin{minipage}[c]{0.49\linewidth}

 \centering 
    \includegraphics[width=1\linewidth]{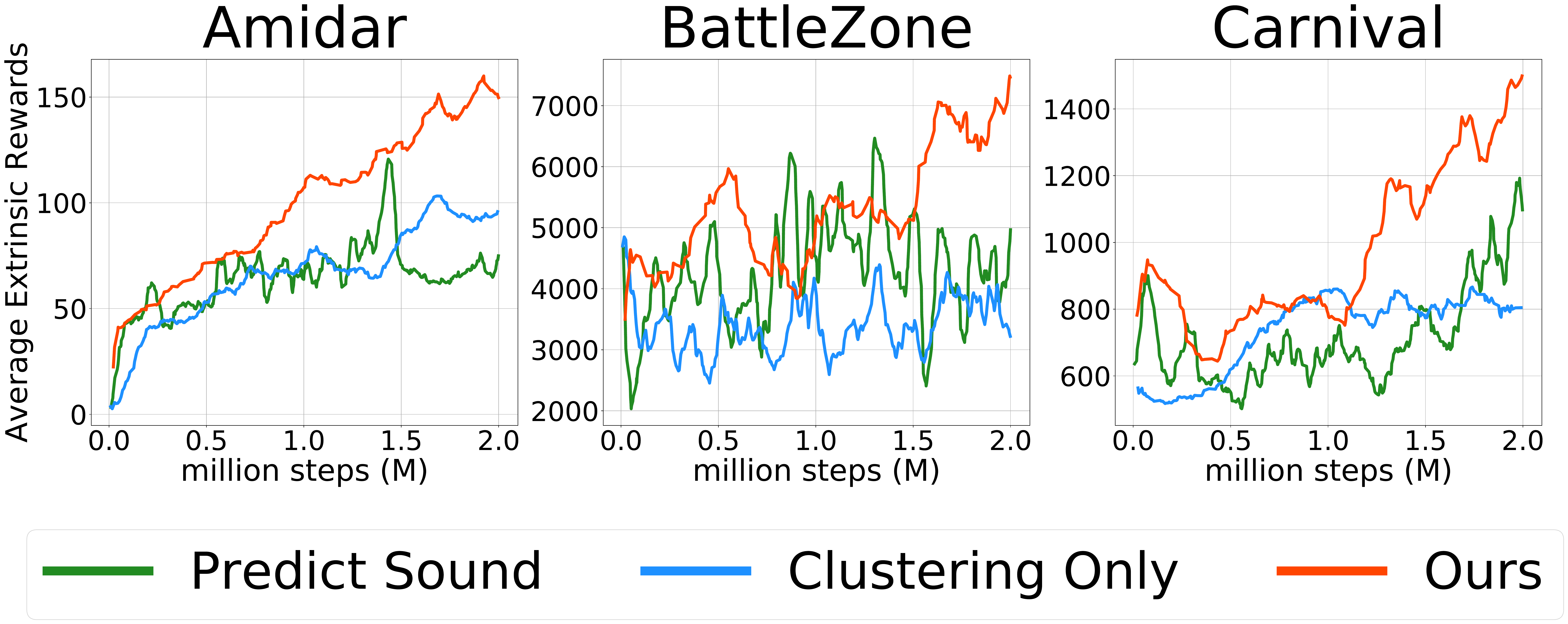}
    
    \caption{Comparisons on earned extrinsic rewards between our auditory event prediction module and sound feature prediction module.}
    \label{fig:event}
\end{minipage}
 \hfill
 \begin{minipage}[c]{0.49\linewidth}
 
 \centering 
    \includegraphics[width=1\linewidth]{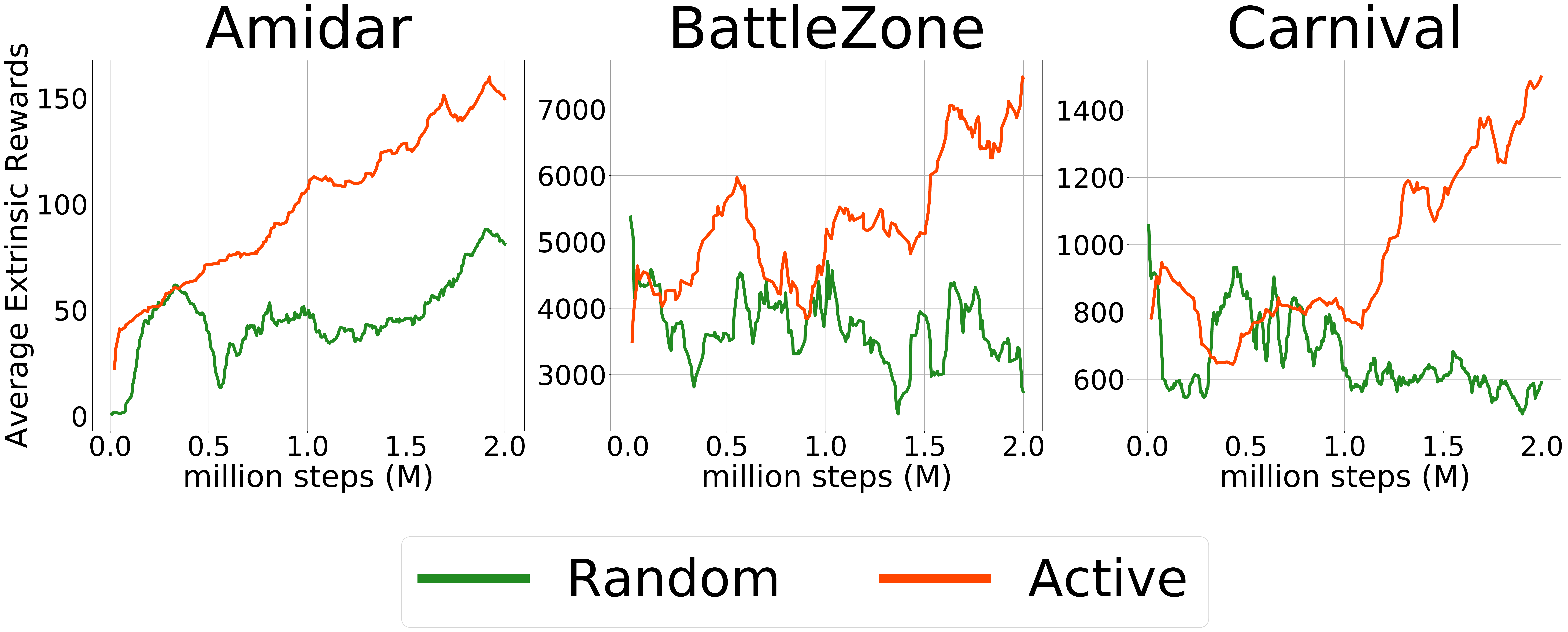}
   
    \caption{Comparisons on earned extrinsic rewards between our active exploration and a random exploration strategy.}
    \label{fig:active}
\end{minipage}

\end{figure*}

%% file: supp.tex
\newpage

\appendix

\begin{center}
{
\LARGE \textbf{Supplementary Material}
}
\end{center}

We start by providing an in-depth understanding of our algorithm works well under what circumstances in Section~\ref{sec:result_analysis}.  We then provide the details of hyper-parameters used for training RL algorithm in Section~\ref{sec:training}. Lastly, we then conduct an ablated study on audio signals in Section~\ref{sec:ablated}.

\section{Result Analysis}
\label{sec:result_analysis}
In this section, we would like to provide an in-depth understanding of our algorithm works well under what circumstances. The sound effects in Atari games fall into three different categories: 1) event-driven sounds which emitted when agents achieve a specific condition (\textit{e.g.}, picking up a coin, the explosion of an aircraft, etc.); 2) action-driven sounds which emitted when agents implement a specific action (\textit{e.g.}, shooting, jumping, etc.) and 3) background noise/music.  According to the dominant sound effects in each game, we summarize the 20 Atari games in Table \ref{table:games}.

\begin{table}[h]
\centering
\caption{Category results of 20 Atari games according to the dominant type of sound effects. We label the games in which our method performs the best in \textbf{bold front}.}
\vspace{7mm}
\label{table:games}
\begin{tabular}{c|c}
\hline
Dominant sound effects & Atari games \\ \hline
Event-driven sounds & \begin{tabular}[c]{@{}c@{}}\textbf{Amidar}, \textbf{Carnival}, \textbf{NameThisGame}, \textbf{Frostbite}, \\ \textbf{FishingDerby}, \textbf{MsPacman} \end{tabular}                        \\ \hline
Action-driven sounds   & \begin{tabular}[c]{@{}c@{}}\textbf{AirRaid}, \textbf{Assault}, \textbf{Jamesbond}, \textbf{ChopperCommand}, \textbf{StarGunner}, \\   \textbf{Tutankham}, \textbf{WizardOfWor}, \textbf{Gopher}, DemonAttack \end{tabular} \\ \hline
Background sounds    & Asteroids, Freeway, TimePilot, BattleZone, CrazyClimber                                                                                                        \\ \hline
\end{tabular}
\end{table}

Based on the category defined in Table~\ref{table:games} and the performance shown in Figure 3 in the main paper, we can draw three conclusions.  First, both event-driven and action-driven sounds boost the performance of our algorithm. Since the sound is more observable effects of action and events, understanding these casual effects is essential to learn a better exploration policy. Second, our algorithm performs better on the games dominant with event-driven sounds compare with those with action-driven sounds. We believe that event-driven sounds contain higher-level information, such as the explosion of an aircraft or collecting coins, which can be more beneficial for the agent to understand the physical world. Third, our algorithm falls short in comparison with baselines when the sounds effects mainly consist of meaningless background noise or background music (\textit{i.e.} CrazyClimber, MsPacman, Gopher, Tutankham, WizardOfWor, and BattleZone). These sounds have little relevance to visual clues and cannot provide useful information or rewards to agents.

\section{Training Details}
\label{sec:training}
Table \ref{table:hyp} shows the hyper-parameters used in our algorithm.

\begin{table}[h]
\centering
\caption{Hyper-parameters used in our algorithms.}
\vspace{7mm}
\label{table:hyp}
 	\begin{tabular}{c|c}
		\toprule 
		Hyperparameter & Value \\ 
		\midrule 
		Rollout length & 128 \\
		Number of minibatches & 4 \\
		Learning rate  & 2.5e-4 \\
		Clip parameter & 0.1 \\
		Entropy coefficient & 0.01 \\
		$\lambda$ & 0.95 \\
		$\gamma$ & 0.99 \\
		\bottomrule 
	
\end{tabular}
\end{table}

\clearpage
\section{Ablated Study}
\label{sec:ablated}
In this section, we carry out ablated experiments to demonstrate that the gains in our method are caused by the audio-event prediction, rather than the use of multi-modality information. For four baselines (\textit{i.e.} RND, RFN, ICM, and DIS), instead of predicting audio-event, they consider sound information by concatenating both visual and sound features to predict the image embedding in the next time step. As shown in Figure~\ref{fig:sup_ab}, our algorithm significantly outperforms other baselines in five Atari games. This indicates that it is non-trivial to exploit sound information for RL, and our algorithm benefits from the carefully designed audio-event prediction as an intrinsic reward.

\input{sup_ab}

%% file: sup_ab.tex
\begin{figure*}
    \centering 
    \includegraphics[width=1\linewidth]{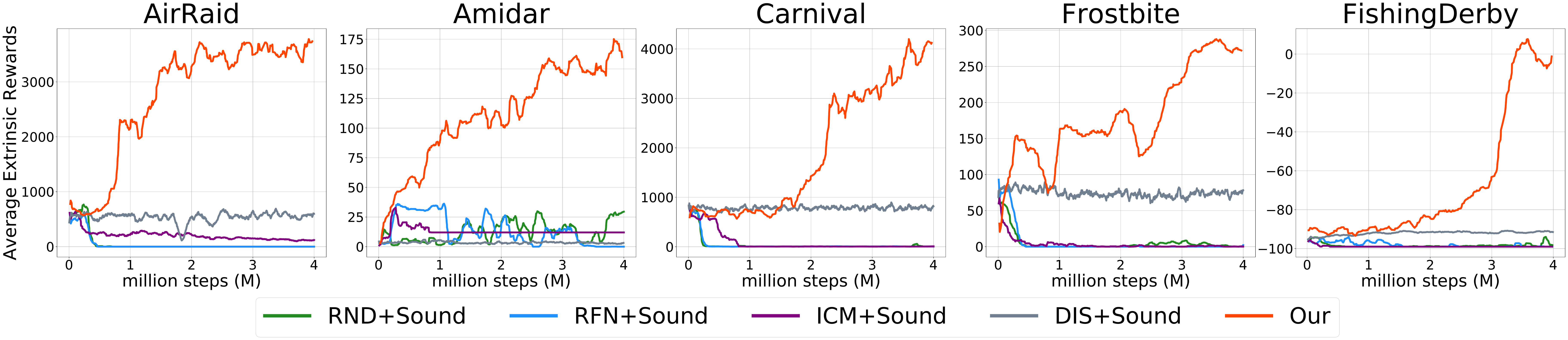}
  
    \caption{Average extrinsic rewards of our model against baselines combined with sound in 5 Atari games}
    \label{fig:sup_ab}
\end{figure*}